\documentclass[twocolumn]{article}
\usepackage{mathptmx}
\usepackage{amsmath}
\usepackage{graphicx}
\usepackage{bm}

\begin{document}
\twocolumn[
\begin{center}
{\bf
{\LARGE Extraction of the principal skeleton of a character as a graph from a character image}

Kazuhisa Fujita

University of Electro-Communications

1-5-1 Chofugaoka, Chofu, Tokyo, 182-8585, Japan

k-z@nerve.pc.uec.ac.jp
}
\end{center}
]

\begin{abstract}
    This paper aims to make a graph representing an essential skeleton of a character from an image that includes a machine printed or a handwritten character using growing neural gas (GNG) method and relative network graph (RNG) algorithm.
    The visual system in our brain can recognize printed characters and handwritten characters easily, robustly, and precisely.
    How does our brain robustly recognize characters?
    The visual processing in our brain uses the essential features of an object, such as crosses and corners.
    These features will be helpful for character recognition by a computer.
    However, extraction of the features is difficult.
    If the skeleton of a character is represented as a graph, we can more easily extract the features. 
    To extract the skeleton of a character as a graph from an image, this paper proposes the new approach using GNG and RNG algorithm.
    I achieved to extract skeleton graphs from images including distorted, noisy, and handwritten characters.
\end{abstract}

\textbf{\emph{Keywords:}} {\it Skeletonization, Character recognition, Self organizing map}

\section{Introduction}

Why can we robustly recognize characters from rotated, distorted, and noisy images including characters?
This ability is provided a robust visual recognition mechanism in the brain.
In the visual processing in the brain, the principal features of an object are used for recognition.
The principal features of an object are such as crosses, corners, circles, and so on.
In pattern recognition by a computer, we may achieve to provide more robust image recognition if we effectively use these principal features.
This study aims to extract principal structures of a character from an image as a graph, which we call a skeleton graph, to efficiently use principal structures of a character for character recognition by a computer.
A skeleton graph represents a skeleton of a character.
Each skeleton graph extracted from the images including the same characters will be similar.
Thus, using the similarity of each skeleton graph allows us to achieve more robust character recognition.
In this study, we propose the method of extraction of a skeleton graph of a character from an image including a character in order to achieve more robust character recognition using similarity of structures of skeletons.

Extraction of a skeleton from a character image is called skeletonization.
Skeletonization is generally excused before the recognition process by a learning machine \cite{Jia:2013}.
Skeletonization is a general morphological method that is used to thin a broad stroke of a character image and to extract only a bone of a character from a character image.
The significant functions of skeletonization in image processing are reducing data size and making more straightforward extract morphological features.
This method allows us to extract a skeleton ``image'' of a character from an image.

To achieve to extract a skeleton graph, we employed growing neural gas (GNG) method that is topology learning algorithm and one of self organizing map (SOM) methods.
SOM \cite{Kohonen:1990} can be developed topology conserving classifiers.
However, the network structure of SOM is static (generally, n-dimensional lattice) and the network structure cannot represent the topology of input space.
Growing neural gas (GNG) method improves this problem because the GNG flexibly increases or decreases nodes and edges of the network.
GNG method has been proposed by Fritzke \cite{Fritzke:1995}.
GNG method have been widely applied to clustering or topology learning, such as reconstruct 3D models \cite{Holdstein:2008}, landmark extraction \cite{Fatemizadeh:2003}, and object tracking \cite{Frezza-Buet:2008}.
We applied GNG method to skeletonization.

In the present study, we demonstrated making a skeleton graph of a character using our proposed method.
Under noisy circumstance, our approach could also produce satisfactory result.
This achievement may allow us to robustly extract a skeleton from a character images.

\section{Methods}
\subsection{Scheme}

\begin{figure}[htbp]
    \begin{center}
    \includegraphics[width=0.3\textwidth]{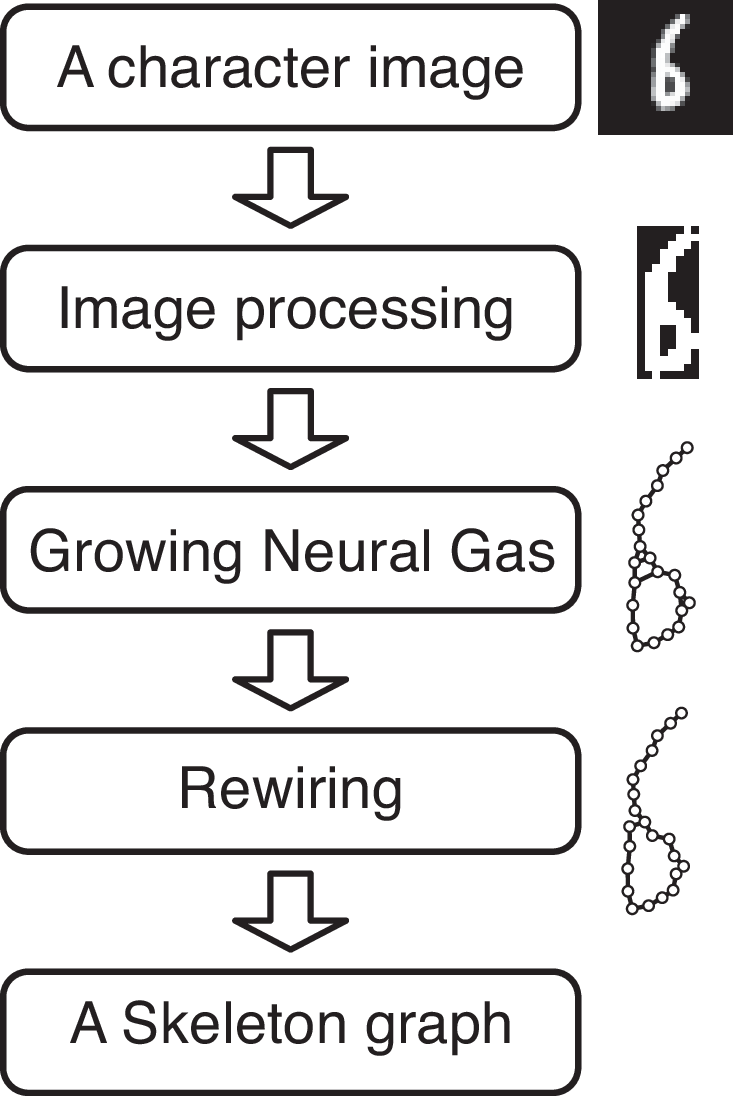}
    \vspace{-.5\baselineskip}
    \caption{Scheme to make a skeleton graph.}
    \label{fig:scheme}
    \end{center}
\end{figure}

Figure \ref{fig:scheme} shows the scheme to make a skeleton graph from an image including a character.
To make the skeleton graph, we use three steps.
The first step, execute image processing that is binarizing and trimming.
The second step, roughly extract a skeleton graph from a character image using the GNG method.
The third step, remove redundant edges and rewire nodes using relative neighborhood graph (RNG) algorithm \cite{Toussaint:1980}.
GNG method is high ability to extract topological features of characters and easily method to assemble. 
However, the graph generated by GNG method has a few redundant edges because GNG method tends to make triangle cluster  \cite{Datta:2001,Garcia-Rodriguez:2012}.
To resolve this problem, we used RNG method that has ability to extract a perceptually meaningful structure.
Using this method, redundant edges reduce and fundamental structures are extracted from a character image.
Through the three steps, the skeleton graph can represent principal structure of a character.

\vspace{-1.5\baselineskip}

\subsection{Growing neural gas}

The growing neural gas (GNG) method have been proposed by Fritzke \cite{Fritzke:1995}.
The GNG method is one kind of SOM methods and extracts topology or classification from data.
The network of GNG flexibly varies and its structure represents data structure.
Using these features, we extracted the skeleton of a character from a character image as a graph.

The input space of the GNG network is an image that is a two dimensional pixel space sized $W \times H$.
The network consists of a set $A$ of nodes.
Each node $c \in A$ has an associated reference vector $\bm{w}_c$.
Node's reference vector $\bm{w}_c$ is denoted by $\bm{w}_c = (w_{cx}, w_{cy})$.
The two parameters represent the node position over the image.
The reference vectors must fulfill:
\begin{equation}
    0 \leq w_{cx} < W,~ 0 \leq w_{cy} < H.
\end{equation}
There are edges between pairs of nods.
These connections are not weighted and not directed.
The edges defined topological structure of the network.

\begin{figure}[htbp]
    \begin{center}
    \includegraphics[width=0.4\textwidth]{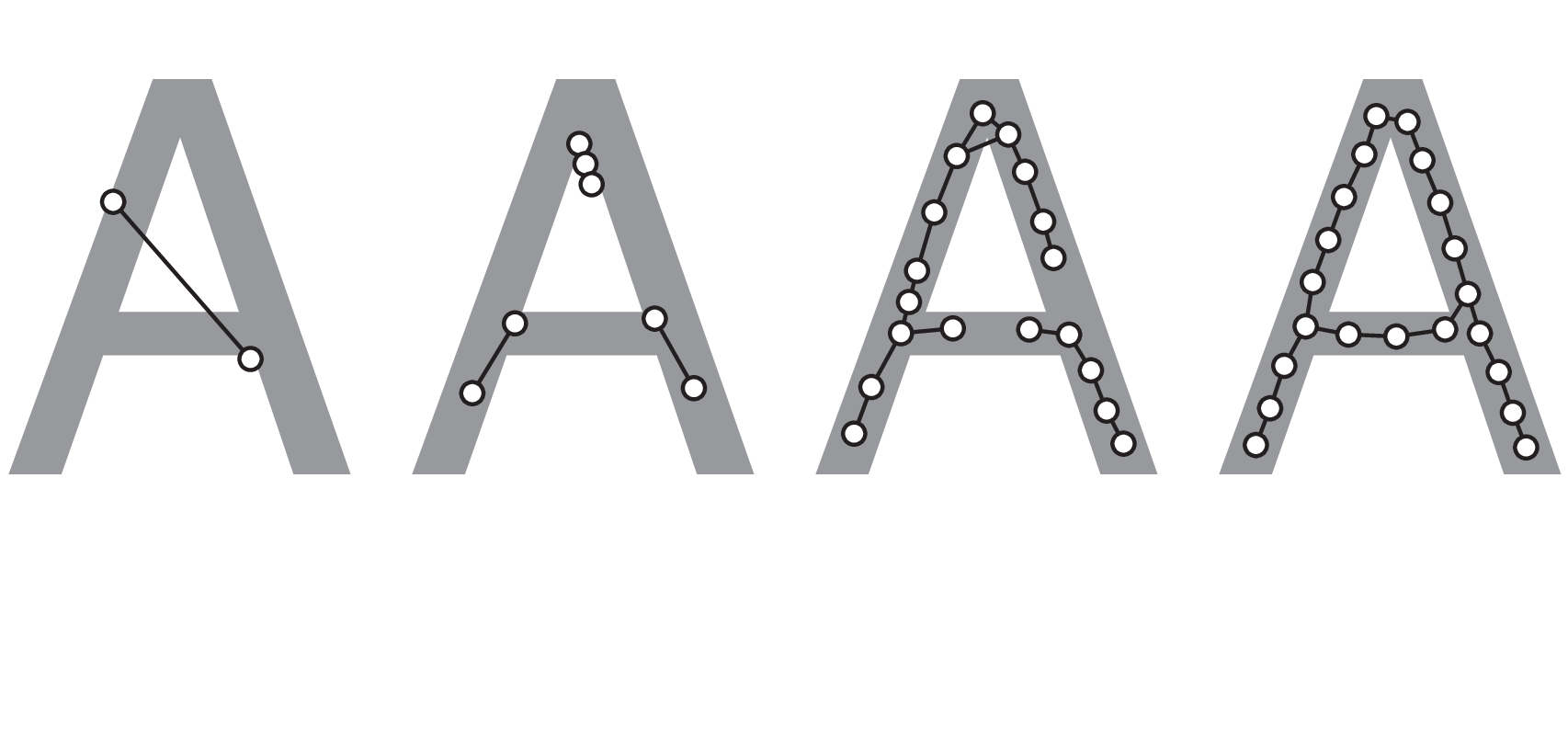}
    \vspace{-.5\baselineskip}
    \caption{Different steps of convergence of the network for the character ``A''. These figures show the networks after 0, 4000, 12000, and 80000 steps (from left to right). At the end of the adaptation process the connection between the nodes represents the topological structure of ``A''.}
    \label{fig:gng}
    \end{center}
\end{figure}

Every node is examined to calculate which one's weight vector are most like the input vector through the following process.
\begin{enumerate}
    \item Starting with only two nodes that are connected each other.
        Positions of the nodes is random in {\bf R}.
    \item Input vector $\bm{x}_i = (x_i, y_i)$ is chosen at random from pixels on a character.
    \item The criterion for neighborhood is Euclidean distance between vectors of an input and a weight of a node.
        The number $k$ of the winning (nearest) node is defined by,
        \begin{equation}
            k = \arg \min_i \| \bm{w}_i - \bm{x}\|.
        \end{equation}
        Simultaneously, find the second nearest node $s$.
    \item Increase the age of all the edges connecting with the wining node.
    \item Add the squared distance between the input vector and the winner node to a local counter variable:
        \begin{equation}
            \Delta \mathrm{error}_k = \|\bm{w}_k - \bm{x}\|^2.
        \end{equation}
    \item The winning node $k$ is rewarded with becoming more like the input vector.
        \begin{equation}
            \bm{w}_k(t + 1) = \bm{w}_k(t) + \lambda(t) (\bm{x} - \bm{w}_{k}).
        \end{equation}
        All direct neighbors $n$ of $k$ are also rewarded.
        \begin{equation}
            \bm{w}_n(t + 1) = \bm{w}_n(t) + \lambda(t) (\bm{x} - \bm{w}_{n}),
        \end{equation}
        where $t$ is learning frequency, $\lambda$ is the learning coefficient.
        $\lambda$ decays with learning frequency.
        \begin{equation}
            \lambda(t) = \lambda_0 \times ( 1 - \frac{t}{T}).
        \end{equation}
    \item If $k$ and $s$ are connected, set the age of this edge to zero. If $k$ and $s$ are not connected, add the edge between these nodes.
    \item Remove the edges with age larger than $a_{\rm{max}}$. If the node isolated by this remove process, remove the node.
    \item Every certain number of input signals generated, insert a new node:
        \begin{itemize}
            \item Determine the neuron $q$ with the maximum summed error.
            \item If the summed error is Error$_0$, insert new node $r$ between $q$ and $q$'s most further neighbor $f$:
                \begin{equation}
                    \bm{w}_r = (\bm{w}_q + \bm{w}_f)/2.
                \end{equation}
                The number of nodes has limit $N_{\rm{max}}$.
                Error$_0$ and $N_{\rm{max}}$ are required to hardly make redundant nodes and edges, and triangle cycles.
        \end{itemize}
    \item If a stopping criterion is not fulfilled, set all error variables to zero and go to step 2.
\end{enumerate}
Figure \ref{fig:gng} shows growing process of the skeleton graph generated by GNG method.
The character image includes ``A''.
It can be seen that GNG network learns the skeleton topology of the character.

The parameters for this simulation were: $\lambda_0 = 0.2$, $N_{\rm{max}} = 40$, $a_{\rm{max}} = 28$.

\subsection{Rewiring}

\begin{figure}[htbp]
    \begin{center}
    \includegraphics[width=0.35\textwidth]{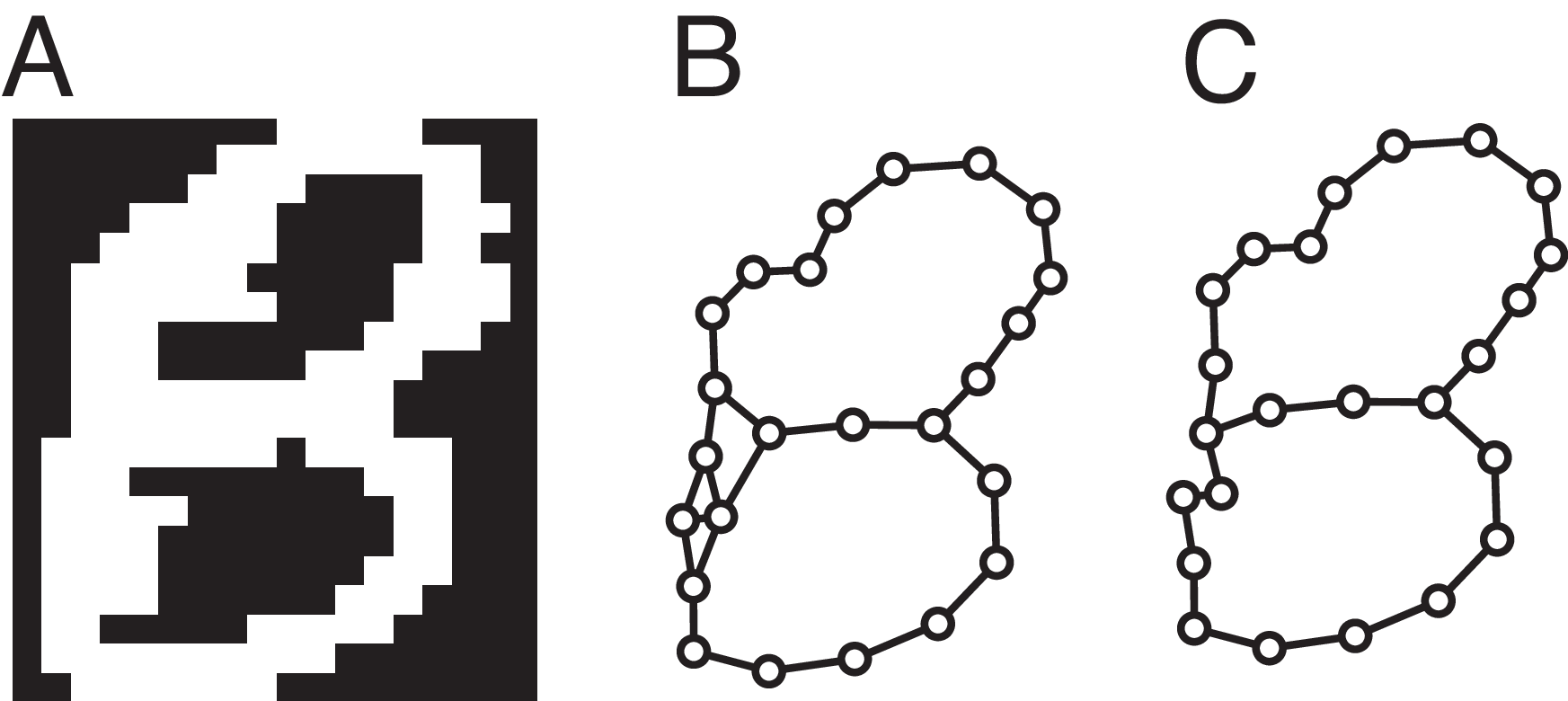}
    \vspace{-.5\baselineskip}
    \caption{A shows the handwritten digit ``8''. B and C illustrate a graph generated by GNG algorithm and rewired graph, respectively.}
    \label{fig:rewiring}
    \end{center}
\end{figure}

The graph generated by GNG method tends to have triangle cycle \cite{Datta:2001,Garcia-Rodriguez:2012}.
The skeleton graph generated by GNG method, shown in fig. \ref{fig:rewiring} B, has the triangle cycles and the redundant edges.
The triangle cycles and the redundant edges especially appeared on which stroke was crossed and on a broad curve line.
To represent principal bone as a graph, deleting the triangle cycles and redundant edges of the skeleton graph generated by GNG method are required.
To reduce the redundant edges, we implemented the rewiring process.
We kept nonredundant edges and deleted redundant edges using Relative neighbor graph (RNG) algorithm \cite{Rocha:1995,Toussaint:1980}.
On RNG algorithm, each node of the network is relative neighbors if the they are near.
If nodes $i$ and $j$ are relative neighbors, there dose not exist another node $z$ of the set such that,
\begin{equation}
    d(z, i) < d(i, j) \text{~and~} d(z, j) < d(i, j),
\end{equation}
where $d(i, j)$ is the Euclidean distance between $i$ and $j$.
When node $i, j$ fulfill the equation except $d(i, j) > \sqrt{W^2 + H^2} \times 0.15$, nodes are connected.
Figure \ref{fig:rewiring} C shows the skeleton graph processed by RNG algorithm.
Redundant edges reduced and principal skeleton graph extracted.

\section{Results}

\begin{figure*}[htbp]
    \begin{center}
    \includegraphics[width=0.9\textwidth]{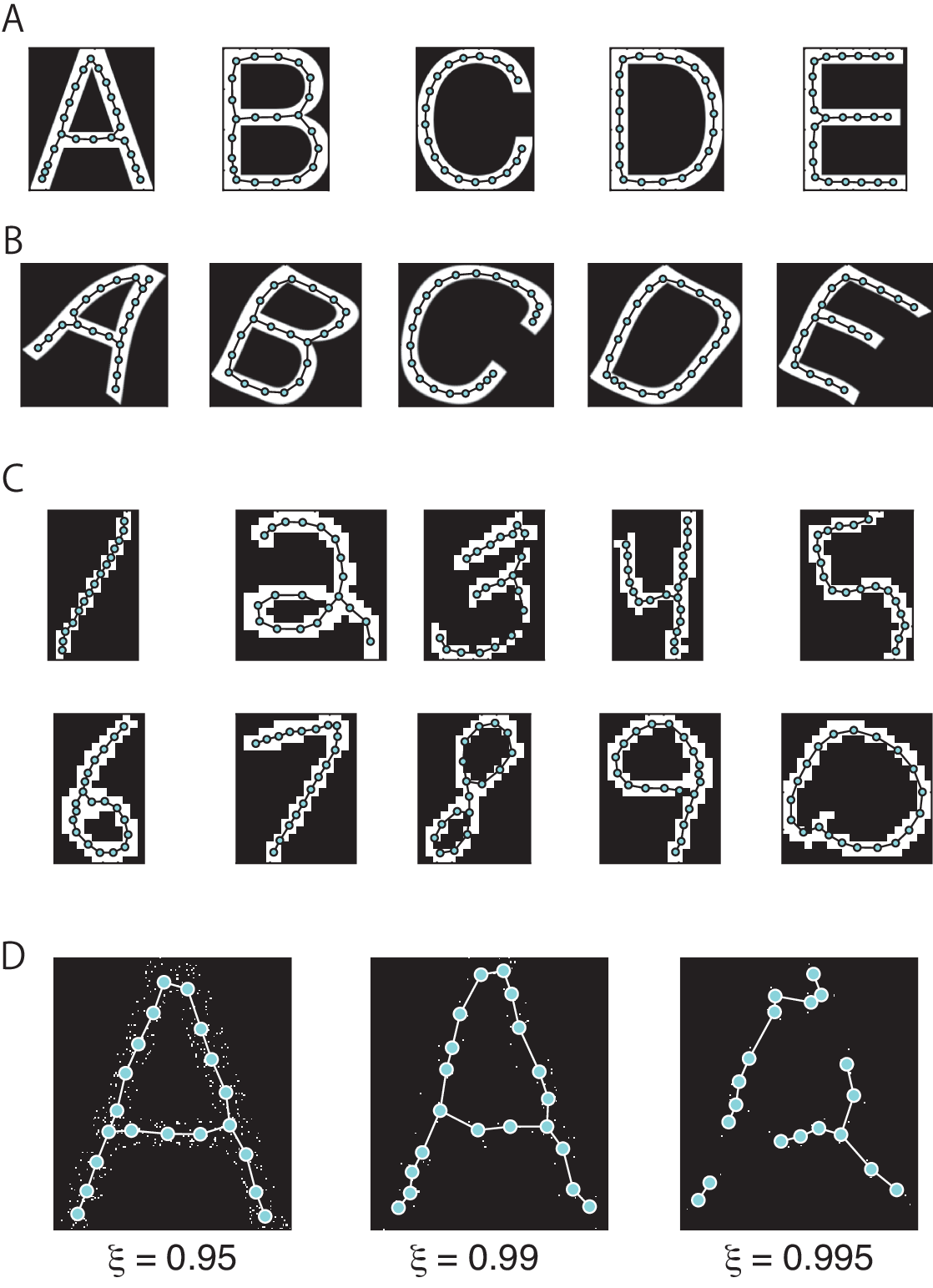}
    \vspace{-.5\baselineskip}
    \caption{(A) Alphabets. (B) Distorted and Rotated alphabets. (C) Handwritten digits. (D) Noised alphabets.}
    \label{fig:skeleton}
    \end{center}
\end{figure*}

To verify that the extracted skeleton graph represents primary structures of a character, the proposed method has been tested with the four sets of characters that are represented by binary images.
The results for the four sets of images are shown in fig. \ref{fig:skeleton}.
The first set consisted of regular images that include undistorted printed-characters.
The skeleton graphs represented the principal structures of the characters, shown in \ref{fig:skeleton} A.
The second set consisted the images of distorted and rotated printed-character.
In this case, the skeleton graphs also represented principal structures.
The topology of the skeleton graphs is almost the same structure as the skeleton graph generated from the regular images.
The third set consisted of the images of isolated handwritten digits from the MNIST Database \cite{Lecun:1998}.
In this case, the skeleton graphs also represented principal structures.

The fourth set consisted of the images of noised characters.
In this test, we randomly changed white pixels on a character to black pixels.
The random noise is uniformly distributed on a character.
Here, we define noise rate $\xi = \nu / \rho$, where $\nu$ is the amount of changed pixels and $\rho$ is the original number of pixels on a character.
Figure \ref{fig:skeleton} D shows the skeleton graphs at different noise levels.
The skeleton graphs produced by our method was consistent with visual form of characters for $\xi = 0.95$ and 0.99.
However, for $\xi = 0.995$, the skeleton graph could not represent the form of characters.

\begin{figure}[htbp]
    \begin{center}
    \includegraphics[width=0.45\textwidth]{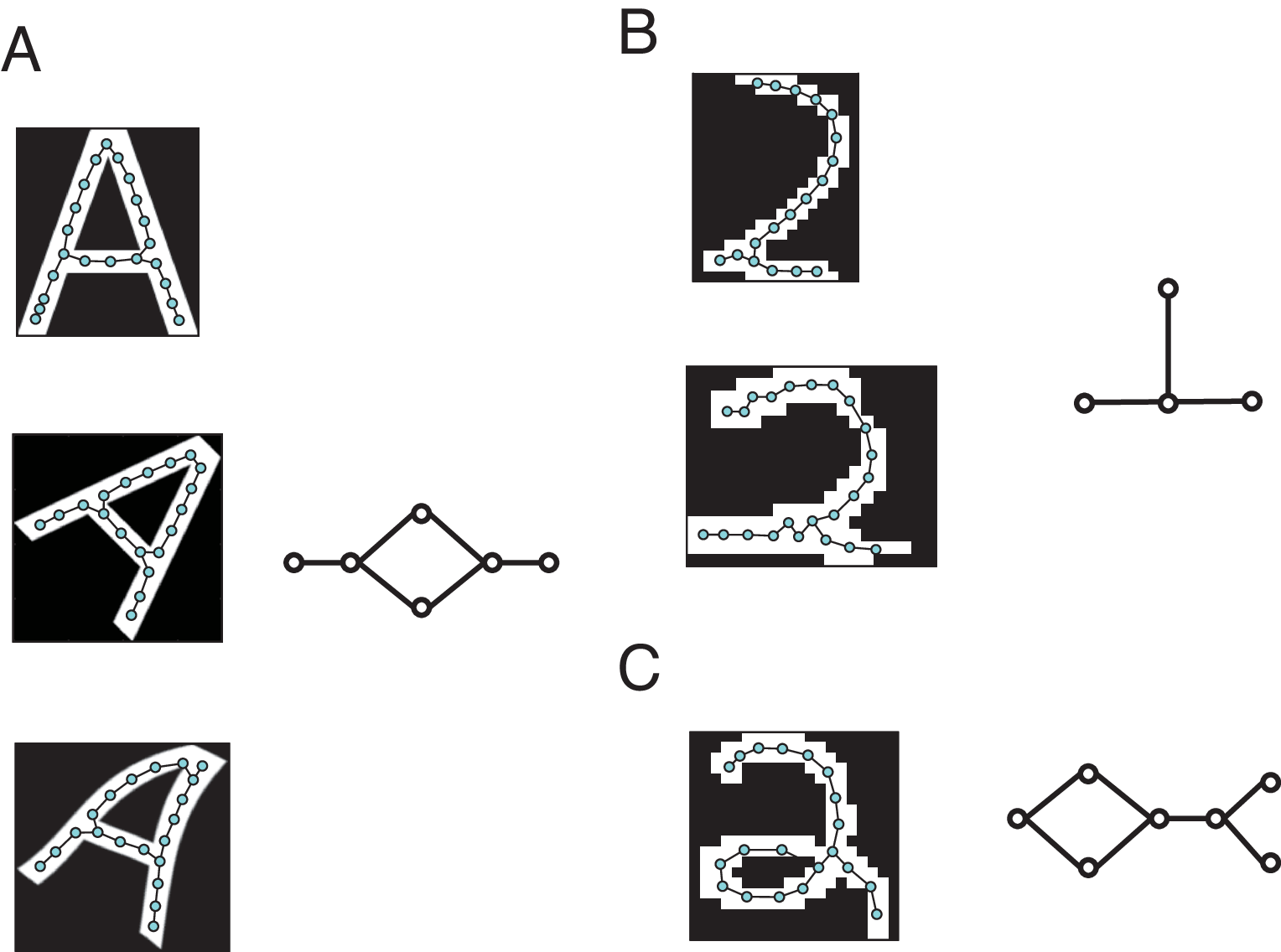}
    \vspace{-.5\baselineskip}
    \caption{Relation between skeleton graphs and topology of the graphs. In the cases of A and B, the skeletons generated form images of same characters had same topology. In the case of C, the skeleton had different topology form one in B.}
    \label{fig:topology}
    \end{center}
\end{figure}

Using our method, we could extract structures of a character as a skeleton graph.
It is important that the skeleton graphs made from images including same character have the common principal skeleton.
The structure of ``A'' has the features that are two T-junctions, one cycle, and one sharp curve. 
If our method effectively extracts skeleton graphs from various ``A'' images, the skeleton graphs must have these features.
Figure \ref{fig:topology} A shows the skeleton graphs extracted from not-distorted ``A'', rotated one, and rotated and distorted one.
These skeleton graphs had the common principal features.

However, skeleton graphs generated from handwritten character images or more distorted character images that include a same character may not always have same features.
For example, the skeleton graphs generated from the handwritten digits ``2'' shown in fig. \ref{fig:topology} B was different from one shown in fig. \ref{fig:topology} C in spite of same digit.
The skeleton graphs shown in fig. \ref{fig:topology} B was the typical skeleton of ``2''.
The skeleton graph generated from a printed character image will also have the same structure.
The typical structure of the skeleton graph of ``2'' is one T-junction.
While the skeleton graph shown in fig. \ref{fig:topology} C was not typical because the graph did not have the typical feature that was one T-junction.
The skeleton graph of fig. \ref{fig:skeleton} C had one junction and one cycle.
These results suggest that the skeleton graphs generated from images including the same character may have different features.

\section{Conclusion and Future Work}

In this paper, we proposed a method to generate a skeleton graph representing the principal features of a character in an image.
We generated a skeleton graph from a character image using GNG method, and then we deleted redundant edges of the skeleton graph using RNG algorithm.
The proposed method has been tested on images including a printed character, a distorted printed-character, a handwritten digit, and a noised character.
The experimental results show the effectiveness of the proposed method.
The skeleton graph preserved an approximation of the original shapes and had principal features of a character.
The topology of the skeleton graph generated by our method did not depend on rotation and distortion of a printed character.

However, skeleton graphs generated from images including same character did not always have same features, for example handwritten characters.
Furthermore, the skeleton graphs generated from images including different characters may have the same topology.
For example, the skeleton graphs made from ``e'' and ``p'' have one cycle and one T-junction, and the topology of these graphs is the same.
In this case, character recognition is not achieved using the topology of the skeleton graphs.
In order to achieve character recognition, the location of nodes and the number of nodes will be required.

In this study, we made a skeleton graph from a binary images.
However, our method can be applied to directly making a skeleton graph from a gray scale image itself.
To extract a skeleton graph from a gray scale image itself, the probability of selection of pixels on a character depends on the intensity of pixels in GNG process.

In future work, we shall develop a character classification method using similarity of skeleton graphs because the skeleton graphs extracted from images including the same character have similar features.

\end{document}